\documentclass[10pt,twocolumn,letterpaper]{article}

\usepackage{wacv}
\usepackage{times}
\usepackage{epsfig}
\usepackage{graphicx}
\usepackage{amsmath}
\usepackage{amssymb}
\usepackage[font=small,labelfont=bf]{caption}
\usepackage[T1]{fontenc}
\usepackage[utf8]{inputenc}
\usepackage{authblk}
\usepackage{graphics}
\usepackage{amssymb}

\wacvfinalcopy 

\def\wacvPaperID{1259} 

\ifwacvfinal\fi
\begin{document}

\title{Learning to Detect Head Movement in Unconstrained \\
Remote Gaze Estimation in the Wild}


\author[1,2]{Zhecan Wang\thanks{olinzhecanwang@gmail.com}}
\author[3]{Jian Zhao*\thanks{zhaojian90@u.nus.edu, https://zhaoj9014.github.io/}}
\author[2]{Cheng Lu*\thanks{lvcheng27@hotmail.com}}{}
\author[2]{Han Huang\thanks{huangh@xiaopeng.com}}
\author[2]{Fan Yang\thanks{fyang@temple.edu}}
\author[2]{Lianji Li\thanks{lilj2@xiaopeng.com}}
\author[2]{Yandong Guo\thanks{yandong.guo@live.com, Corresponding author}}

\makeatletter
\renewcommand\AB@affilsepx{, \protect\Affilfont}
\makeatother

\affil[1]{Columbia University}
\affil[2]{XPENG Motors}
\affil[3]{Institute of North Electronic Equipment, Beijing, China}


\maketitle
\ifwacvfinal\thispagestyle{empty}\fi


\begin{abstract}
\vspace{-4mm}

Unconstrained remote gaze estimation remains challenging mostly due to its vulnerability to the large variability in head-pose. Prior solutions struggle to maintain reliable accuracy in unconstrained remote gaze tracking. Among them, appearance-based solutions demonstrate tremendous potential in improving gaze accuracy. However, existing works still suffer from head movement and are not robust enough to handle real-world scenarios. Especially most of them study gaze estimation under controlled scenarios where the collected datasets often cover limited ranges of both head-pose and gaze which introduces further bias. In this paper, we propose novel end-to-end appearance-based gaze estimation methods that could more robustly incorporate different levels of head-pose representations into gaze estimation. Our method could generalize to real-world scenarios with low image quality, different lightings and scenarios where direct head-pose information is not available. To better demonstrate the advantage of our methods, we further propose a new benchmark dataset with the most rich distribution of head-gaze combination reflecting real-world scenarios. Extensive evaluations on several public datasets and our own dataset demonstrate that our method consistently outperforms the state-of-the-art by a significant margin. 
\end{abstract}

\vspace{-5mm}

\section{Introduction}

\vspace{-2mm}

\begin{figure}[t]
\vspace{-4mm}
\begin{center}
\scriptsize
\scalebox{0.8}{
  \includegraphics[width=1\linewidth]{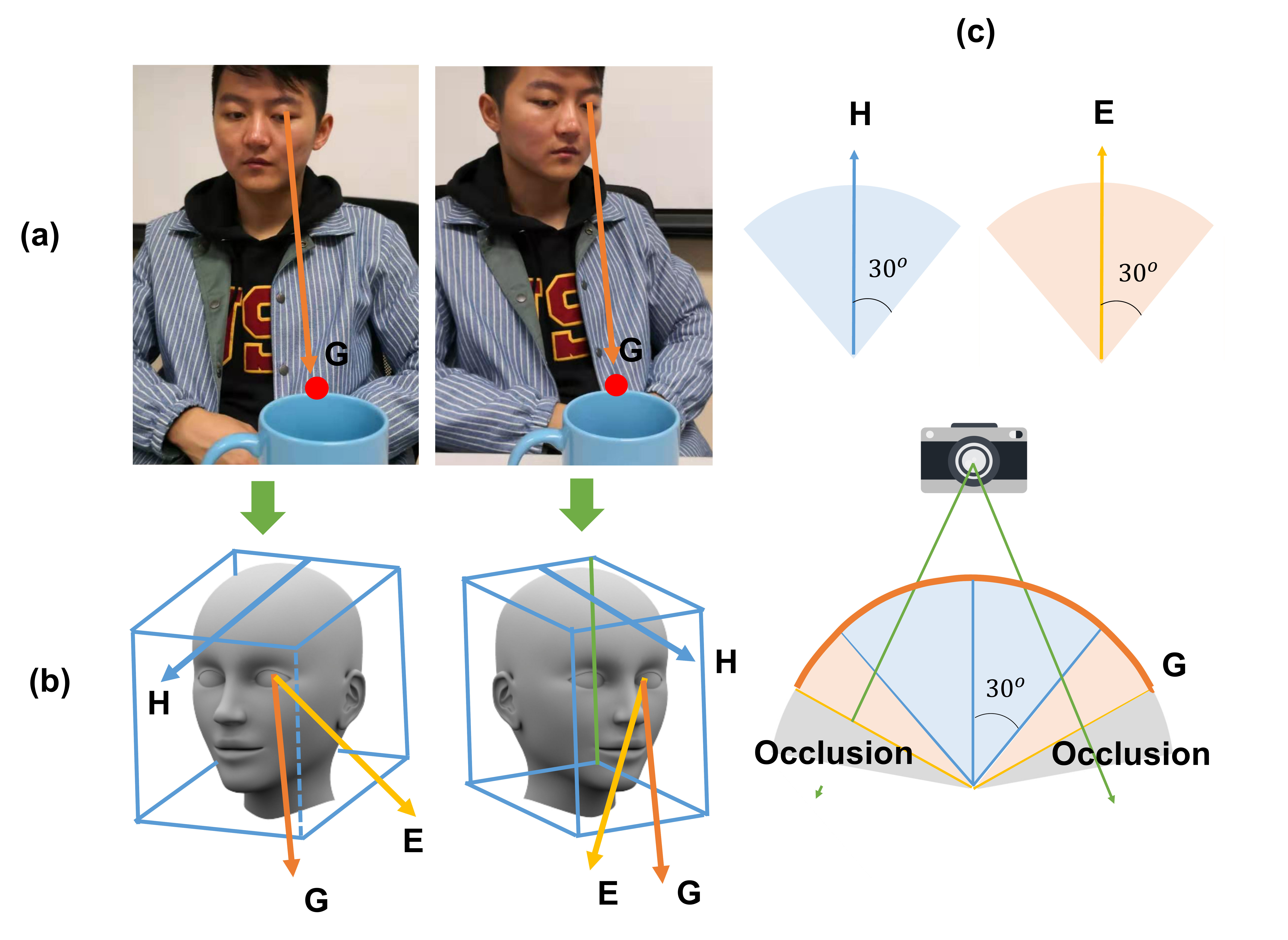}
}
\end{center}
\vspace{-7mm}
  \caption{Effect of head movement on gaze. H represents head-pose vector, E represents eye-ball vector and G represents gaze vector. In (a), without head-pose, both poses map to the same gaze ground-truth respective to camera causing confusion. Even though the gaze vector, G, relative to camera coordinate stays the same, both head-pose vector, H, and eye-ball vector, E, change. However, with head-pose, it is easier to learn the difference and more accurate mapping function to estimate gaze direction. Head movement would also affect gaze distribution \cite{ranjan2018light}. Since eye-ball vector rotates around a given head-pose vector, a function of the observed head-pose \cite{ranjan2018light} is normally the mean of the gaze distribution. Further, as illustrated in (b), assuming the ranges of head movement and eyeball movement are up to 60 degrees. Thus, the head-pose could cover up to $60^{\circ}$ in total. However, based on head-pose, gaze could cover up to $120^{\circ}$ in total. In addition, if the head moves to the edge of its distribution, the eye movement may have occlusions against the camera. These occlusions would cause differences in gaze's ranges of distributions corresponding to head-pose.}
\label{fig:1}
\vspace{-7mm}
\end{figure}

Unconstrained remote gaze estimation has many important applications \cite{meena2018toward, putra2018development, article, richmond2019image, bala2018method, d2012gaze} mostly around {\bf H}uman {\bf C}omputer {\bf I}nteraction ({\bf HCI}) \cite{kumar2007eyepoint, morgante2012critical, vinciarelli2009social}. A variety of existing methods \cite{wang2019generalizing, huang2017tabletgaze, park2018learning, sugano2017s} could achieve very high accuracy in detecting gaze directions under controlled settings.

However, existing methods \cite{lu2014adaptive, huang2017tabletgaze, park2018learning} still suffer from problems like: inaccuracy under real-world conditions, need of complex settings to adapt to free-head movement, low image quality \cite{zhang2017mpiigaze}, offset from personal calibration, \textit{etc}. Among them, head movement perhaps is the most crucial factor that significantly affects unconstrained remote gaze estimation of the following reasons, 1) any gaze vector related to a fixed camera coordinate depends on both eye vector (visual axis of an eyeball \cite{guestrin2006general}) and head-pose vector, 2), as illustrated in Fig. \ref{fig:1}, head-pose also strongly affects gaze distribution including both mean and range \cite{ranjan2018light}, 3) head-pose would change eye appearance \cite{article}. The difference in head-pose would cause geometric deformation. Eye regions like pupil, iris, sclera, \textit{etc}, would be occluded to different extents \cite{ranjan2018light}. Because this deformation is holistic throughout the face, it is too diminutive in a local eye region for appearance-based methods to detect and track especially without personal calibration. With this understanding, we believe gaze estimation could be more robust to the change of eye appearance caused by head movement by incorporating head-pose information. In this work, we introduce two ways of incorporating head-pose into gaze estimation to achieve better accuracy.



Among unconstrained remote gaze estimation methods, appearance-based methods recently become popular due to their general applicability to multiple scenarios \cite{zhang2015appearance, park2018learning, huang2017tabletgaze, krafka2016eye, sugano2017s, zhang2017mpiigaze}. However, they are also not sensitive enough to free-head movement especially when eye ball’s relative position to camera coordinate is fixed, as in Fig. \ref{fig:1}. Furthermore, they are trained and evaluated on public datasets mostly collected under controlled scenarios with very limited illuminations, subject identities, backgrounds, \textit{etc} \cite{fischer2018rt, huang2017tabletgaze, smith2013gaze, weidenbacher2007comprehensive, sugano2014learning, park2018learning, funeyediapes2014, villanueva2013hybrid, mcmurrough2012eye, zhang2017mpiigaze, smith2013gaze}. Most importantly, these datasets lack rich distribution of head-pose. Some of their sampling ranges are even discrete. Due to these problems, these datasets bear the risk of bias and could not generalize to other real-world scenarios, \textit{e.g.} in-car scenarios under sunlight.

We compensate the confusion caused by head movement in gaze estimation by introducing two ways of incorporating head-pose. Our work focuses on proposing a system to incorporate head-pose in two different scenarios. First, when direct head-pose information, \textit{i.e.} facial image and head-pose vector, is available, we propose {\bf H}ead-pose-aware {\bf G}aze {\bf D}etector ({\bf HGD}), an appearance-based method that leverages head-pose and gaze in an end-to-end structure. Different from previous works like \cite{ranjan2018light, zhang2015appearance}, our method merges head and gaze information more properly in different levels of representations including hidden feature level, training task level, and model level. On each level, these representations are merged with similar spatial dimensions and information complexities. HGD outperforms the state-of-the-art in both public datasets and our dataset. Furthermore, in some scenarios (datasets) where direct head-pose information is not preserved, we additionally propose a side method, {\bf HGD}-{\bf no}-{\bf H}ead-{\bf P}ose ({\bf HGD-noHP}), that could also incorporate head-pose into gaze estimation by extracting head-pose information from eye deformations. In order to evaluate our methods better on a benchmark closely reflecting real-world scenarios, we further collect our own dataset, \textit{i.e.} In-car Gaze dataset. In this dataset, we collect data from both head and eye movement over much larger continuous ranges compared with existing datasets.



Our contributions are summarized as follows.
\begin{itemize}
\vspace{-2mm}
  \item We propose an end-to-end method, HGD and one side method, HGD-noHP, for better incorporating head-pose in gaze estimation in the wild. 
  \vspace{-2mm}
   \item For better evaluating our frameworks, we collect a large-scale benchmark with richer head-gaze distribution better reflecting real world scenarios.
   \vspace{-2mm}
  \item Comprehensive evaluations on the In-car Gaze dataset proposed in this work and other existing datasets verify the superiority of our frameworks on gaze estimation in the wild over the state-of-the-art.
\end{itemize}

\begin{figure*}[hbt]
\vspace{-4mm}
\begin{center}
\scalebox{0.9}{
  \includegraphics[width=1\linewidth]{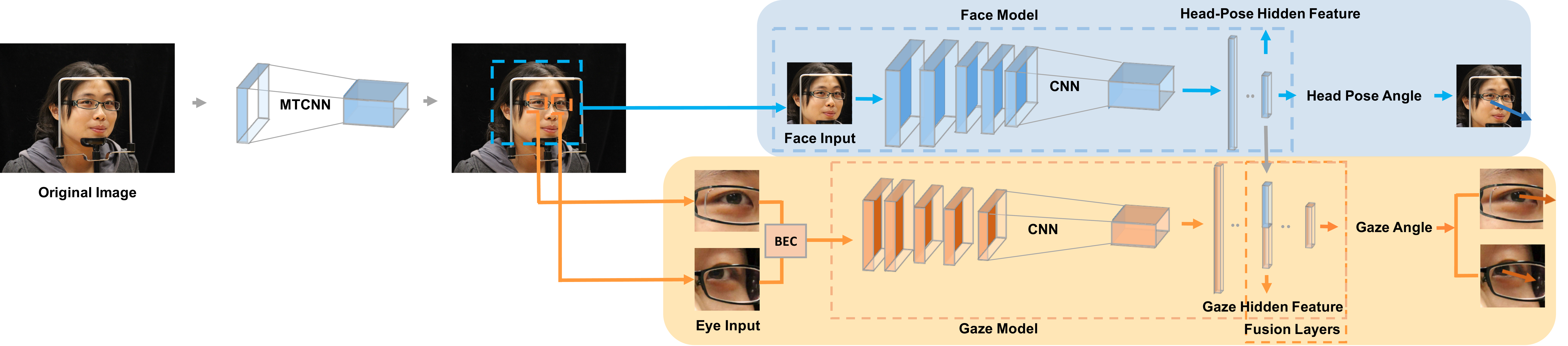}
}
\end{center}
\vspace{-4mm}
  \caption{Structure of HGD. In this framework, head-pose and gaze are merged and have a balance in input level (we use both face and eye images as inputs in the input level for models), hidden feature level (concatenation between two hidden features), model level (parallel relationship between head model and gaze model) and task level (parallel relationship between head-pose task and gaze task). The blue part represents the training and testing on head-pose information and the red part on gaze information (we enhance the resolution of input images here for demonstration purpose). Best viewed in color.}
  
\label{fig:_1}
\vspace{-6mm}
\end{figure*}

\vspace{-4mm}

\section{Related Work}

Recent remote gaze estimation methods focus more on head-free gaze estimation by incorporating head-pose information \cite{wang2019generalizing, fischer2018rt, wang2017real, valenti2011combining, lu2011head}. They could be divided into two main categories, \textit{i.e.} appearance-based and model-based methods. 

Model-based methods often use the geometric prior to regularize models for gaze estimation. They are previously widely explored for good accuracy and ability to handle free-head movement by using multiple light sources or cameras under controlled settings \cite{kar2017review, lai2014hybrid, ohno2004free, beymer2003eye, zhu2007novel}. They could be divided into two parts, {\bf P}upil {\bf C}enter {\bf C}orneal {\bf R}eflection methods ({\bf PCCR}) \cite{article, guestrin2006general} and non-PCCR methods depending on if using external light sources or not. PCCR methods could be precise in controlled scenarios but impractical for real-world scenarios. Non-PCCR methods include 3D model-based methods \cite{meyer2006single, guestrin2006general, hennessey2006single} and 2D shape-based methods \cite{cherif2002adaptive, zhu2002subpixel}. 3D model-based methods and 2D shape-based methods directly infer gaze from observed eye shapes, such as pupil center or iris edges. If applied to real-world scenarios, model-based methods could not easily adapt to free-head movement, low image quality, different lightings or subjects without extra calibration. This complexity limits them from being applied to more general environments.

Appearance-based methods directly use eye images as input and can, therefore, work with low-resolution images. Because they are typically data-driven, they could leverage large amounts of head-pose independent training data to generalize to arbitrary users without extra setup or calibration. Current works using monocular cameras become more attractive given its generality \cite{zhu2017monocular}. Even though existing appearance-based methods do include head-pose information in the pipelines but they do not incorporate it properly. In existing methods like \cite{cheng2018appearance, fischer2018rt, zhang2015appearance, ranjan2018light}, the measured 3D head-pose vector is directly inserted into the second last {\bf F}ully {\bf C}onnected ({\bf FC}) layer. This direct concatenation would be difficult for the last FC layer to learn since the 3D head-pose vector is very different from the learned hidden features in terms of spatial dimension and embedded information complexity. In other ways of incorporating head-pose, \cite{krafka2016eye} proves that a {\bf C}onvolutional {\bf N}eural {\bf N}etwork ({\bf CNN}) that takes multi-region inputs, \textit{i.e.} eyes, faces and face grids, can improve gaze estimation. These regions, mainly the full face, could better encode head-pose, geometric structure of head and illumination across larger areas than those available in the eye region. Thus, from experiments in \cite{krafka2016eye, sugano2017s}, we observe that full-face-based CNNs are more accurate for gaze estimation than eye-image-based CNNs. However, they are limited in applicability as the entire face may not be available in all scenarios \cite{huang2017tabletgaze}. Furthermore, the method proposed in \cite{krafka2016eye} may be limited to 2D-screen scenarios and the full face-based method in \cite{sugano2017s}, only using full face as input, may be more vulnerable to low image quality where eye regions could be more blurry. Unlike \cite{zhang2015appearance, ranjan2018light}, our methods merge head-pose and gaze when they are in similar levels of representations, \textit{e.g.} merging between hidden feature vectors, parallel learning tasks, \textit{etc}. Different from \cite{krafka2016eye, sugano2017s}, our methods are also not limited to 2D screens, less vulnerable to low image quality and could better generalize to different scenarios.



\vspace{-3mm}

\section{Gaze Dataset}

Even though many public gaze datasets are already available \cite{fischer2018rt, huang2017tabletgaze, smith2013gaze, weidenbacher2007comprehensive, sugano2014learning, funeyediapes2014, villanueva2013hybrid, mcmurrough2012eye, zhang2017mpiigaze, smith2013gaze, krafka2016eye}, many of them \cite{mcmurrough2012eye, weidenbacher2007comprehensive, smith2013gaze} are collected in controlled laboratory settings and have limitations in scales, subjects, ranges of sampling, \textit{etc} \cite{sugano2014learning, funeyediapes2014, zhang2015appearance, huang2017tabletgaze}. These limitations would cause problems like lack of variation for subject appearances, head-pose, gaze, \textit{etc} \cite{mcmurrough2012eye, villanueva2013hybrid, weidenbacher2007comprehensive, smith2013gaze}, and further prevent appearance-based methods from better generalization \cite{wang2019generalizing}. Thus, we collect our benchmark, In-car Gaze, closer to real-world scenarios to train more generalized appearance-based methods and more clearly demonstrate the advantage of our frameworks.

In-car Gaze not only has the largest continuous ranges of sampling for gaze and head-pose but also has one of the largest scales in frames. Many of the datasets like \cite{mcmurrough2012eye, villanueva2013hybrid} underplay the collection of head-pose information, \textit{e.g.} most of them do not store facial images but only eye images. This causes an imbalance in the distribution, stored data format and quantity between head-pose and gaze information. We overcome this by focusing on the collection of both head-pose and gaze. Besides, most datasets \cite{sugano2014learning, zhang2015appearance, fischer2018rt} are recorded under controlled scenarios having limited participants \cite{sugano2014learning, funeyediapes2014, zhang2015appearance, huang2017tabletgaze} and environment settings like illumination conditions and backgrounds. Differently, we invite 1000 participants with diverse facial appearances. Furthermore, our dataset is the only dataset that labels both left eye and right eye on the same face respectively with two different gaze ground-truths and has multi-camera views per shot (supplementary, Sec. In-car Gaze Dataset).

\vspace{-2mm}

Our work do not solely focus on car driving scenarios. Different from existing car gaze solutions and datasets \cite{palazzi2018predicting, schwarz2017driveahead, wang2019continuous, naqvi2018deep, vora2018driver}, our frameworks and datasets focus more on improving general gaze estimation by incorporating head-pose information. The flexibility of our solutions and detailed labelling of In-car Gaze dataset could generalize to other daily life scenarios.

\vspace{-2mm}

\section{Proposed Method}

\vspace{-1mm}

In the following sections, we introduce methods to incorporate head-pose into gaze estimation in two different scenarios: head-pose learned from human face when direct head-pose information, \textit{i.e.} both face images and head-pose labels, is available and head-pose learned from eye deformations when direct head-pose information is not available \cite{article}. When merging, we consistently unify head-pose and gaze representations in a similar level of spatial dimension and embedded information complexity. We believe this intuitive strategy would help our models better incorporate head-pose to reach higher gaze estimation accuracy. Furthermore, realizing that the distance between two pupils causing asymmetry, we find that {\bf B}oth {\bf E}yes concatenated on the {\bf C}hannel ({\bf BEC}) level could help achieve the best accuracy and efficiency compared with single eye method and else, referring to {\bf Component Analysis}. Thus in both of our frameworks, eye images are pre-processed in {\bf BEC} method on our dataset but in the fashion of single eye in public datasets when paring information is not available.

\subsection{Head-Pose Learned from Human Face} 


Eye image, the direct local information, is important for gaze prediction. However, for appearance-based methods, the change of eye appearances from head movement may be too diminutive to detect solely from this local information. Thus, the change of eye appearances caused by head-pose would cause confusion for the regressor. To solve this, we introduce extra global information by bringing in full face information. This is because geometric deformation caused by head movement will be more distinctly expressed in the scale of full face. We further formulate this learning problem as a task of learning a transformation function, $F_{transform}$, from eye, $X_{eye}$, and face, $X_{face}$, to gaze prediction, $g_{w}$, as in Eqn. \ref{eq:4}. With this intuition, in scenarios, \textit{e.g.}, our collected dataset, where both facial images and head-pose labels are available, we propose our main method, HGD, as illustrated in Fig. \ref{fig:_1}. The original image is passed through a MTCNN face detector \cite{zhang2016joint} to produce face image and eye images based on detected landmarks. Then the remaining framework learns both head-pose (as the blue part) and gaze (as the red part) from these face and eye images. \cite{sugano2017s} uses spatial weights to focus on the edges, the geometric layout of face besides eye regions. Different from that, in our method, this weighting could be implicitly learned through the head-pose prediction task from face image. 
We use a simple ResNet-34 \cite{he2016deep} structure as the face model to learn head-pose directly from face image, as the top part in Fig. \ref{fig:_1}. In this setting, head-pose information is implicitly embedded in the geometric structure of the provided face image. The face model outputs a $64 \times 1$ feature vector from the second last FC layer and a 3D head-pose angle (yaw, pitch). This part is formulated as the first equation in Eqn. \ref{eq:5}. We then also have a ResNet-34 \cite{he2016deep} as the gaze model to produce a gaze hidden feature ($64 \times 1$) at its FC-3 layer, formulated as the second equation in Eqn. \ref{eq:5}. The gaze model concatenates the shared head-pose feature with its gaze feature and then outputs to the fusion layers, following FC layers, to predict gaze. This part is formulated as the last part in Eqn. \ref{eq:5}, also illustrated in Fig. \ref{fig:_1}. From this framework, the gaze model not only learns head-gaze relationship from the back-propagation from both training tasks but also from the concatenated features. This end-to-end schema allows the model more easier handle low image quality and adapt to different scenarios.

For implementation, depending on the distribution of head-gaze distribution in different datasets, we have two training strategies for this structure.

{\bf Multi-task, Implicit Learning:} Public datasets, as Columbia Gaze \cite{smith2013gaze} or MPII Gaze \cite{zhang2015appearance} datasets shown in Fig. \ref{fig:4}, usually have insufficient combination of head-pose and gaze due to insufficient collection of head-pose. Therefore, it would be easier for the model to learn the head-gaze relationship even though these datasets may not truly reflect the real-world scenarios. In this case, we train the face model and the gaze model jointly on two parallel tasks, one head-pose loss and one gaze loss, as in Eqn. \ref{eq:6}. The learning of face model and the designated loss function would force the gaze model to learn the relationship between gaze and head-pose thus helping gaze prediction. The backpropagation from two losses would simultaneously constrain face model and gaze model mutually. We set the model to mainly learn to predict gaze and assist this learning with an ancillary head-pose task. Purposely, we multiply the head-pose loss with a weakening factor so as to strengthen the gaze learning during training. Because of the intrinsic characteristics of deep learning, we could not fully supervise the whole learning process during multi-task learning and ensure that the gaze model could learn head-gaze relationship properly in every step. Consequently, we call this implicit learning strategy. 

 {\bf Multi-stage, Explicit Learning:} During training in our dataset, we realize that the losses of both head-pose and gaze could not converge jointly as well as we experience in public datasets like Columbia Gaze \cite{smith2013gaze}. This may be due to facts that in real-world scenarios as in our dataset, the distribution of head-gaze is very dispersed, as shown in the right of Fig. \ref{fig:4}. Also, different from most public dataset collected in controlled laboratory scenarios, our dataset is collected in daily scenarios and the labelling could be rough. Thus it would be more difficult for the model to learn the relationship between head-pose and gaze jointly online. In this case, we sacrifice computation efficiency to conduct a multi-stage training strategy. We first only train the face model with the head-pose loss until it converges well. Then we freeze the face model and use its inferenced output, \textit{i.e.} head-pose hidden features, to feed the gaze model for gaze prediction. Under this strategy, we are able to secure the stable performance of both models with less internal constraints during training. In this setting, we specifically train the face model on head-pose prediction and it back-propagates only on its own and so as gaze model. Comparatively, we call this strategy explicit learning since we separate the learning processes explicitly upon two tasks. As shown in Tab. \ref{Table:4} and Tab. \ref{Table:5}, HGD achieves the best accuracy in our dataset and public datasets where head-pose information is well preserved.

\begingroup\makeatletter\def\f@size{8}\check@mathfonts

\vspace{-3mm}

\begin{equation} \label{eq:4}
g_{w} = F_{transform} (X_{eye}, X_{face}),
\end{equation}

\vspace{-5mm}

\begin{equation} \label{eq:5}
 F_{transform}  = \left\{\begin{matrix}
 V_{face} = CNN_{face}(X_{face}, W_{face}), \\ 
 V_{eye} = CNN_{eye}(X_{eye}, W_{eye}),\\ 
 g_{w} = F_{fusion}(V_{face}, V_{eye}),
\end{matrix}\right.
\end{equation}

\vspace{-5mm}

\begin{equation} \label{eq:6}
 Loss_{batch} = \sum_{n = 1}^{M} (Loss_{gaze}^{(n)}(g_{w}, l_{g}) + \beta \cdot Loss_{head}^{(n)}(h, l_{h})).
\end{equation}

\endgroup

Where $F_{transform}$ indicates the predict function from eye and face to gaze, $g_{w}$ represents the predicted gaze, $CNN_{face}$ and $CNN_{eye}$ represent head-pose and gaze model respectively, $V_{face}$ and $V_{eye}$ are respective hidden features, $W_{face}$ and $W_{eye}$ represent parameters in both models respectively, $F_{fusion}$ means the fusion layers, M is the batch size, $\beta$ is the weakening factor,  both $l_{g}$ and $l_{h}$ are the ground-truth labels and $h$ means the head-pose.

\begin{figure}[t]
\begin{center}
  \includegraphics[width=1\linewidth]{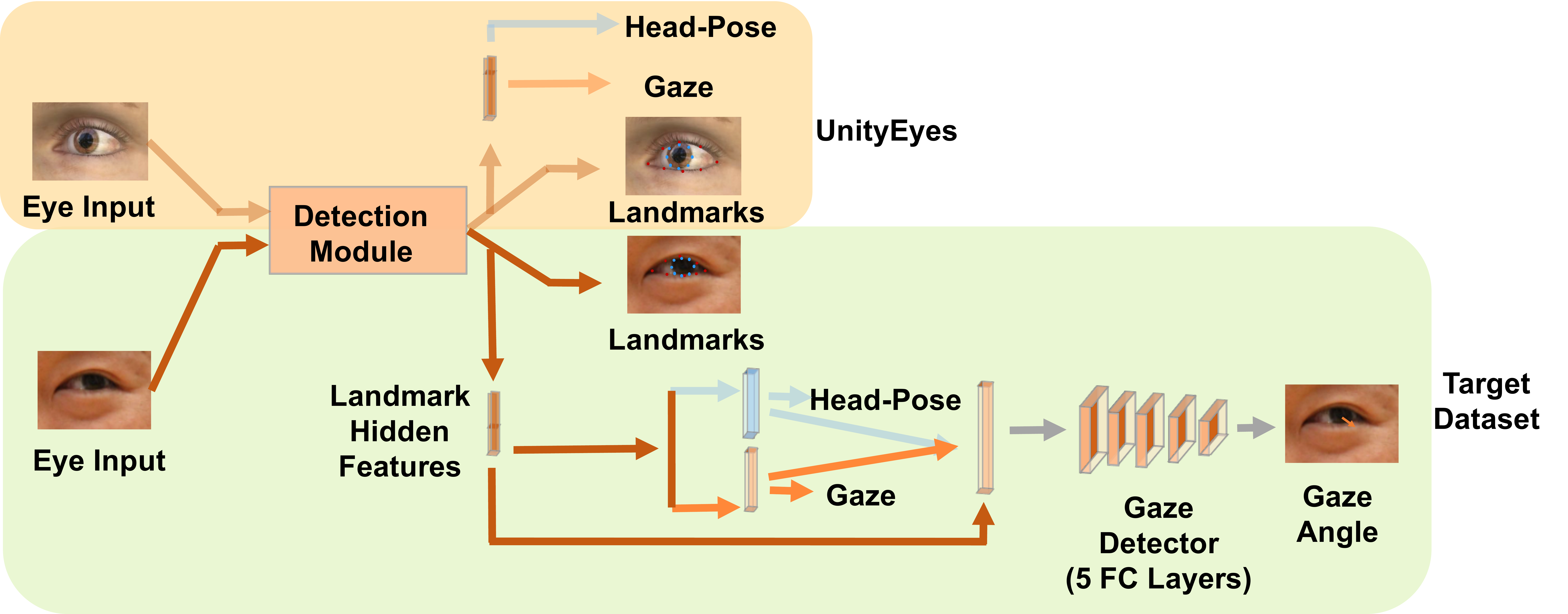}
\end{center}
\vspace{-4mm}
  \caption{Structure of HGD-noHP. The yellow area denotes that the landmark detector is trained on photo-realistic synthetic data from UnityEyes \cite{wood2016learning}. The green area denotes the inference of the landmark detector and the training of gaze detector on target dataset.}
\label{fig:8}
\vspace{-8mm}
\end{figure}

\vspace{-1mm}

\subsection{Head-pose Learned from Eye Deformation}

\vspace{-1mm}

\begin{figure}[t]
\begin{center}
\scriptsize
\scalebox{1}{
  \includegraphics[width=1\linewidth]{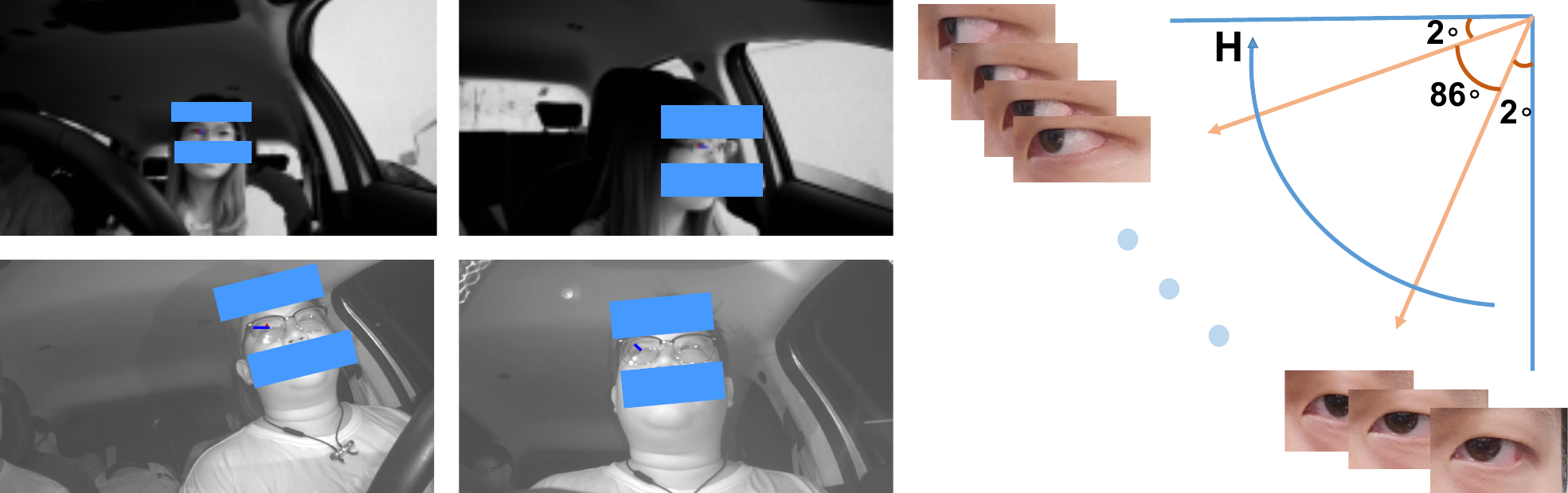}
}
\end{center}
\vspace{-3mm}
  \caption{Example of multi-camera views in car. We collect data through 4 cameras in different perspectives to allow models more generally learn gaze estimation in real-world scenarios. The 1st row shows images collected in daytime scenarios and the 2nd row shows images collected in nighttime. Deformation of eye appearance respective to camera view due to change of head-pose without eyeball movement. The change of appearance may not be clear when head movement is small but obvious when large.}
\label{fig:eye_appearance}
\vspace{-5mm}
\end{figure}

In some of the public datasets, no head-pose information is provided. In real-life scenarios, sometimes only eye images would be provided for remote gaze estimation so our model needs to be very robust to the offset from free-head movement while maintaining accuracy. As mentioned previously, based on works from \cite{article, ranjan2018light}, we know that head movement would change eye appearances, as further demonstrated in Fig. \ref{fig:eye_appearance}. With investigation, we realize that head-pose can also be approximated reversely from eye appearances solely, mainly eye features, \textit{e.g.} shapes of pupils and iris. 

After inspired by model-based gaze estimation algorithms \cite{park2018learning}, we designed a new appearance-based algorithm, HGD-noHP, that focuses on predicting gaze from eye's deformations. Eyeball movement would mainly force the movements of pupils and iris regions causing deformations respective to camera. However head movement would not only cause the deformations from pupils and iris but also the overall structure of eye regions including eyelids, \textit{etc}. It is indeed hard to differentiate between these two kinds of causation relationships explicitly. Thus, instead of directly learning attention maps or gazemaps as in \cite{park2018deep} to mask out specific regions like iris or eyeballs, we utilize labeled data from UnityEyes \footnote{https://www.cl.cam.ac.uk/research/rainbow/projects/unityeyes/tutorial.html} \cite{wood2016learning} to learn those two mappings from two target losses, \textit{i.e.} head-pose and gaze losses. We believe this implicit learning could best utilize the strength of learning based methods. 

\cite{park2018learning} trains a tremendously large hourglass model on synthetic data to predict eye’s landmarks and has a model-based framework followed to estimate the gaze based on these predicted landmarks. Differently, we use a much simpler model, ResNet-34 as the detection module to achieve better computation efficiency, shown in Fig. \ref{fig:8}. Its main task is to serve as the backbone of a landmark detector to predict 16 landmarks of eye’s interior margins and iris (8 landmarks for each category, 32 units in total). We first train the landmark detector on synthetic eye images from UnityEyes. Then we freeze the landmark detector and extract the inferenced hidden features out from the second last FC layer of 200 units. Further, we feed those hidden features to two additional modules, gaze module and head-pose module. Each of them consists of 5 FC layers to train to predict gaze and head-pose respectively on UnityEyes. Those learning tasks would train two modules to learn the important mappings from landmark hidden features to gaze and head-pose. 

Different from \cite{park2018learning} only directly using the predicted landmarks, on the target dataset, \textit{e.g.} Columbia dataset, we first extract inferenced features from landmark detector. Based on that, we also extract inferenced features of 200 units from both gaze and head-pose modules. We concatenate those three hidden features as input to train a final gaze model of 5 FC layers. We believe the concatenated hidden features have richer information about deformations of eye corresponding to both head-pose and gaze than just landmarks. Furthermore, we also use the SimGAN \cite{shrivastava2017learning} trained on the target dataset to help improve the synthetic data by adding more realistic elements (supplementary, Sec. Synthetic Eyes from UnityEyes Improved by SimGAN). This would compensate the accuracy loss due to the decreased capacity of landmark detector. By learning landmark features in the first stage, we add prior knowledge to guide the first part of this framework and believe it would learn the essential geometric features of eyes and the mappings from those geometric deformations to both gaze and head-pose. The training is conducted on photo-realistic synthetic data from UnityEyes improved by SimGAN since labeling eyes’ landmarks to the details of interior margins and iris could be very ambiguous and tedious. 
We further demonstrate that this side framework may not achieve better accuracy than our main proposed method but still outperforms the state-of-the-art in our dataset as shown in Tab. \ref{Table:4} and other existing datasets like Columbia \cite{smith2013gaze} and MPIIGaze \cite{zhang2015appearance}, as in Tab. \ref{Table:5} and Tab. \ref{Table:6}.



\vspace{-3mm}

\section{Experiments}
\vspace{-1mm}

In this section, we first list our implementation details and two evaluation metrics. We then thoroughly analyze the importance of different parts in our algorithms through component analysis. Lastly, we not only evaluate our algorithms with the state-of-the-art on public datasets but also on our own benchmark. 

\subsection{Implementation Details}

{\bf Data Preprocessing:} In our dataset, images are stored in grayscale and sometimes have overexposure due to various lightings. For alleviating the effect of overexposure, we use {\bf M}ultilevel {\bf H}istogram of {\bf O}riented {\bf G}radients ({\bf MHoG}) \cite{dalal2005histograms} + {\bf L}inear {\bf D}iscriminant {\bf A}nalysis ({\bf LDA}) which is invariant to various illuminations to certain extent, as suggested in \cite{huang2017tabletgaze}. We use MHoG + LDA to first extract features from images and then concatenate it with the original images to feed into the framework (supplementary, Sec. 10 Data Preprocessing). This preprocessing could help improve testing accuracy in our dataset, as in Tab. \ref{Table:2}. 

{\bf Training:} We leverage Pytorch \cite{paszke2017automatic} as the implementation environment and our experiments are conducted on a single NVIDIA GPU with 16 GB memory. Our frameworks are trained for 100 epochs with batch size of 64. The input images are set to be $224 \times 224$. The starting learning rate is set to 0.0001 and decays by 0.1 every 30 epochs. Wing loss \cite{feng2018wing} is adopted in our methods. For HGD, after many experiments, the weakening factor, ${\beta}$ is empirically set to 0.3 during online multi-task training. In HGD-noHP, gaze and head-pose module each consists of 5 FC layers of sizes: 200, 200, 100, 50 and 2. The gaze model consist of 5 FC layers of sizes: 600, 300, 100, 32 and 2 to predict gaze. We use UnityEyes to generate 100,000 synthetic images (90,000 for training, 10,000 for testing). 

{\bf Evaluation Metrics:} Different papers use their own evaluation metrics as in \cite{kar2018performance}. For sharing the same evaulation standard, we consistently use two methods in our work (supplementary, Sec. Evaluation Metrics). {\bf V}ector {\bf E}rror {\bf M}etric ({\bf VEM}) calculates the 3D angle difference between the predicted, ${P}$, and the labeled 3D vector, ${R}$, as in Eqn. \ref{eq:1} and Eqn. \ref{eq:2}. We also use {\bf A}ngle {\bf E}rror {\bf M}etric ({\bf AEM}) to calculate the real difference in angular values between the predicted angle, $(\Theta_{p}, \alpha _{p})$, and labeled angle, $(\Theta_{r}, \alpha _{r})$, and ensure their real values are not far off, as in Eqn. \ref{eq:3}. 

\begingroup\makeatletter\def\f@size{8}\check@mathfonts

\vspace{-3mm}

\begin{equation} \label{eq:1}
{P} = T(\Theta_{p}, \alpha _{p}),  {R} = T(\Theta_{r}, \alpha _{r}),
\end{equation}

\vspace{-3mm}

\begin{equation} \label{eq:2}
D_{VEM} = \arccos ({P}\cdot {R}),   
\end{equation}

\vspace{-3mm}

\begin{equation} \label{eq:3}
D_{AEM} = \frac{1}{2n}\sum \left | \Theta_{p} - \Theta_{r}  \right | + \left | \alpha_{p} - \alpha_{r}  \right |. 
\end{equation}

\endgroup

Where T represents the transform function from 3D angles to vectors, ${P}$ represents the predicted angle, ${R}$ represents the labeled angle and $n$ represents the number of samples in test data.

\vspace{-2mm}

\subsection{Evaluation on Public Dataset}

\vspace{-1mm}

\subsubsection{Evaluation of Gaze Estimation with Direct Head-pose Information}

\vspace{-2mm}

To better demonstrate the advantage of our algorithms over the state-of-the-art, we further evaluate our algorithms over three public datasets.

\begin{table}[hbt]
\begin{center}
\centering
\scriptsize
\resizebox{\columnwidth}{!}{
\begin{tabular}{|c|p{2.6cm}|c|c|c|c|c|}
\hline
Backbone                   & Framework  & \multicolumn{2}{|c|}{Columbia \cite{smith2013gaze}} & \multicolumn{2}{|c|}{MPII \cite{zhang2015appearance}} & GazeCapture \cite{krafka2016eye}\\ \hline
\multicolumn{2}{|c|}{} & AEM & VEM & AEM & VEM & MSE         \\ \hline

& HGD - Exp &  0.84 &  {\bf 1.32} & NA   & NA   &   {\bf 2.10}  \\ \cline{2-7}

        {ResNet-34 \cite{he2016deep} }           & HGD - Imp &  {\bf 0.82} &  1.35 & NA   & NA   &  NA       \\ \cline{2-7} 
          & HGD-noHP &  1.94  & 3.32  & {\bf 4.02} & {\bf 5.33} & 3.39  \\ \cline{2-7} 
          & MPIIGaze \cite{zhang2015appearance} &  5.42 & 8.02 & 4.41 & 6.38 & 6.93      \\ \cline{2-7}                 
                            \hline
  & HGD - Exp & 1.52 & 2.49 & NA   & NA   &    2.49     \\ \cline{2-7} 
          {Lenet \cite{lecun1998gradient}}                & HGD - Imp  & 1.59 & 2.41    & NA   & NA   & NA        \\ \cline{2-7} 
          & HGD-noHP & 2.34 & 3.45  & 4.31 & 5.52 & 3.92  \\ \cline{2-7} 
                         & MPIIGaze \cite{zhang2015appearance}                       & 5.32 & 8.26   & 4.51 & 6.43  & 8.03        \\ \hline
                           
                          & iTracker \cite{krafka2016eye}                      & 4.1 & 7.32 & NA   & NA   & 2.13        \\ \hline
                                                 &  RedFTAdap \cite{DBLP:journals/corr/abs-1904-10638}                      & 3.54 & NA & 5.35 & NA   & NA        \\ \hline 
                                                 
                                                 &PictorialGaze \cite{park2018deep}
                         & 3.8 & NA & 4.5 & NA   & NA        \\ \hline
         &  Bayes-adversarial \cite{wang2019generalizing}                      & NA & NA & 4.3 & NA   &  NA   \\ \hline
\end{tabular}
}
\end{center}
\vspace{-4mm}
\caption{ Comparison of our algorithms with the state-of-the-art on public datasets (cross-subject). Eye image input is pre-processed in the fashion of  single eye per unit.}
\label{Table:5}
\vspace{-2mm}
\end{table}


As in Tab. \ref{Table:5}, in all three public datasets, our frameworks could outperform the state-of-the-art. For a fair comparison, we also replace the backbone of our frameworks with Lenet \cite{lecun1998gradient} and they still achieve better accuracy than the state-of-the-art. MPII Gaze dataset \cite{zhang2015appearance} (not MPIIFaceGaze \cite{sugano2017s}) does not provide facial images and is collected in front of laptops causing limited distributions of head-gaze combination. However, HGD-noHP could still take benefit from incorporating head-pose related information to outperform or achieve a comparable accuracy against the state-of-the-art. Even though GazeCapture \cite{krafka2016eye} is collected using phones or tablets and has a smaller distribution of head-gaze, our frameworks could still generalize on it. With this constraint, our frameworks may not be able to significantly demonstrate its advantage in incorporating head-pose information for gaze estimation. However, they could still achieve better accuracy against the state-of-the-art. 

\vspace{-4mm}

\subsubsection{Evaluation of Gaze Estimation without Direct Head-pose Information}

\vspace{-1mm}

As mentioned earlier, in scenarios where direct head-pose information is not available through vector format or facial images, we could infer head-pose information through geometric deformations from eye. Our HGD-noHP framework focuses on learning eye features first and then transfer to gaze prediction. In Tab. \ref{Table:6}, HGD-noHP outperforms the state-of-the-art by a significant margin with head-pose information removed purposely. This signifies the strong relationship between eye features and head-pose. 

\vspace{-2mm}

\begin{table}[hbt]
\begin{center}
\centering
\scriptsize
\begin{tabular}{|c|c|c|c|}
\hline
                                                                   &     & Columbia \cite{smith2013gaze} & UnityEyes \cite{wood2016learning} \\ \hline
{HGD-noHP} & AEM & {\bf 1.94}     & {\bf 2.34 }  \\ 
                          & VEM & {\bf 3.32}     & {\bf 3.45} \\ \hline
{HGD-noHP w/o SimGAN \cite{shrivastava2017learning}}  & AEM & 2.51     & NA         \\ 
                      & VEM & 4.17     & NA \\ \hline
{ResNet-34 \cite{he2016deep}}            & AEM & 3.29      & 4.24                                  \\ 
              & VEM & 5.39     & 5.92    \\ \hline
{MPIIGaze \cite{zhang2015appearance}}                           & AEM & 5.4      & 5.12                  \\ 
                                                                      & VEM & 8.42     & 7.98      \\ \hline
{M-3D Gaze \cite{zhu2017monocular}}  & AEM & 4.09     & 4.87  \\ 
                              & VEM & 6.2      & 5.67      \\ \hline
\end{tabular}
\end{center}
\vspace{-4mm}
\caption{ Comparison of HGD-noHP and other algorithms on public dataset when head-pose information is removed purposely (degree).}
\label{Table:6}
\vspace{-3mm}
\end{table}

\vspace{-5mm}

\begin{table}[hbt]
\begin{center}
\centering
\footnotesize
\scalebox{0.8}{
\begin{tabular}{|l|l|l|}
\hline
 Method  & AEM & VEM        \\ \hline
HGD - Imp & 3.69 & 5.17 \\ 

HGD - Exp w/ BEC & {\bf 1.79}  & {\bf 2.87}\\ 

HGD - Exp  & 2.53  & 3.67\\ 

HGD-noHP w/ BEC & 2.94  & 4.6 \\

HGD-noHP     & 3.21  & 4.97 \\ 

iTracker    \cite{krafka2016eye}                   & 5.61  & 8.64 \\ 

iTracker with ResNet-34       & 5.56  & 8.07 \\ 

MPIIGaze \cite{zhang2015appearance}                       & 4.49  & 6.61 \\ 

MPIIGaze with ResNet-34         & 3.71 & 5.44 \\ 
M-3D Gaze \cite{zhu2017monocular} with ResNet-34       & 5.7   & 9.57 \\ \hline
\end{tabular}
}
\end{center}
\vspace{-3mm}
\caption{ Comparison of different head-gaze merging algorithms on our dataset (degree).}
\label{Table:4}
\vspace{-5mm}
\end{table}


\subsection{Evaluation on the Real-World In-car Gaze Dataset}

During driving, the driver has a relatively broader range for head-pose and gaze among daily life activities, so we select driving as our base scenarios for data collection. As demonstrated in Fig. \ref{fig:4} (more detailed comparison in Tab. 1 of supplementary), In-car Gaze have the largest continuous sampling ranges of head-pose and gaze compared with existing datasets. 1,000 participants are invited from all different kinds of age groups and body traits to ensure the diversity. The collection is conducted inside a car with window and sunroof glasses open sitting outdoors throughout daylight and night to imitate the real-life daily scenarios. For designing a robust system, participants are also asked to wear a variety of different attires including sunglasses, glasses, hats, \textit{etc}. Different from most, we also preserve facial images and label the gaze ground-truths for both left eye and right eye independently from the same face. Last but not least, 400 images are captured for each participant and a large scale of 400,000 frames are stored. 4 near infrared cameras (better visibility, less noise at night than RGB cameras) are set up inside the car in different positions toward the driver. During collection, our machine navigates a laser pointer point to the front within a prefixed grid. For each point, 4 photos are produced from 4 sync cameras, as in Fig. \ref{fig:eye_appearance}. In our dataset, we also store 9 facial landmarks, eye patches, face patches, recovered gaze ground-truths of both left and right eyes, and head-pose vectors. 

  

\begin{figure}[t]
\begin{center}
\scalebox{1}{
  \includegraphics[width=1\linewidth]{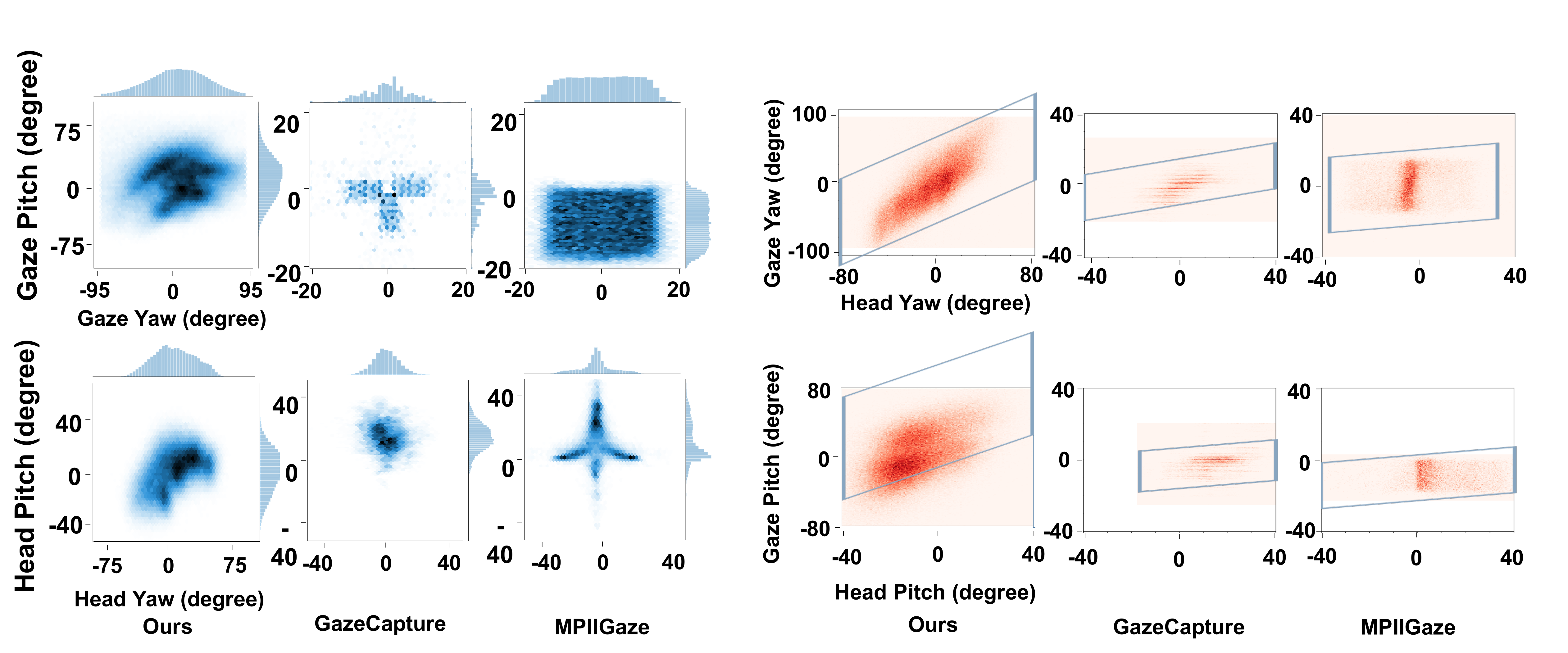}
}
\end{center}
\vspace{-6mm}
  \caption{
  Comparison of distribution of head-pose and gaze across 3 datasets. On the left image, the first row shows the distribution of gaze and the second one shows the distribution of head-pose. The right image shows the distribution of head-gaze combination. The first row demonstrates the distribution in yaw direction and the second one in pitch. Our dataset has the biggest sampling ranges for both head-pose and gaze in both directions.
}
\label{fig:4}
\vspace{-6mm}
\end{figure}



We, in depth, compare our methods of head-gaze merging with the state-of-the-art. In order to evaluate the algorithms fairly, we also replace the backbone of iTracker \cite{krafka2016eye}, MPIIGaze \cite{zhang2015appearance} and M-3D Gaze \cite{zhu2017monocular} frameworks with ResNet-34 \cite{he2016deep}. As in Tab. \ref{Table:4}, all of our presented algorithms could achieve better accuracy than the state-of-the-art with a significant margin. By merging head-pose representation, HGD with explicit learning method achieves the best accuracy. HGD-noHP, due to their limit in head-pose representation, achieve slightly worse accuracy. The original iTracker framework \cite{krafka2016eye} takes left eyes, right eyes, faces and face grids as inputs for the gaze estimation task with 2D on-screen settings. However, in more general real-life settings, face grid may not be necessarily related with gaze estimation but causes more noises. Furthermore, MPIIGaze \cite{zhang2015appearance} and M-3D Gaze \cite{zhu2017monocular} do not merge head-pose and gaze as comprehensive as our frameworks, either, thus having accuracy drop in their corresponding results. 


{\bf Classification:} The practical use of this work is to assist to detects driver's attention. Thus, it may not be necessary to fully determine the exact angle where the driver is looking. In this case, we split the frontal space of the driver into 9 sub-spaces (categories), modify HGD structure to a classifier and plot its results as in Fig. \ref{fig:10} (supplementary, Sec. Regression and Classification). From the confusion matrix, we could see that HGD could accurately catch most of the gaze actions in practice.

\subsection{Component Analysis}

{\bf Significance of Head-pose Information:} In Tab. \ref{Table:2}, the first row represents our main proposed method, HGD. It demonstrates that when all components are included, HGD framework could achieve the best accuracy. From top to bottom, we in sequence get rid of MHoG + LDA, head-pose task and face model. 
As a result, we observe increasing gaze errors which demonstrate the importance of head-pose information in gaze estimation comparatively. Additionally, as listed in Tab. \ref{Table:_depth}, incorporating head-pose information into gaze estimation could consistently gain improvements across different backbones. Our work focuses on proposing a novel framework for incorporating head-pose into gaze estimation under two different scenarios regardless of backbones. Our solution is general to various backbone neural networks including Lenet, ResNet-34, ResNet-52, ResNet-101, ResNet-121, \textit{etc}.

\vspace{-2mm}

\begin{table}[hbt]
\begin{center}
\centering
\footnotesize
\resizebox{\columnwidth}{!}{%
\begin{tabular}{|c|c|c|c|c|c|c|}
\hline
\multicolumn{1}{|l|}{} & Face Model & Head-Pose Task & mHoG + LDA & AEM  & VEM  \\ \hline

{}   & \checkmark             & \checkmark     & \checkmark  & {\bf 1.79} & {\bf 2.87}\\ 
{HGD}   & \checkmark             & \checkmark   & x     & 2.33  & 3.53 \\ 
{} & \checkmark             & x      & x    & 3.95 & 6.25\\ 
{}      & x              & x       & x    & 6.88 & 8.48 \\ \hline
\end{tabular}
}
\end{center}
\vspace{-5mm}
\caption{Comparison of HGD with various components on In-car Gaze dataset.}
\vspace{-2mm}
\label{Table:2}
\end{table}

\vspace{-3mm}

\begin{table}[hbt]
\vspace{-2mm}
\begin{center}
\centering
\scriptsize
\resizebox{\columnwidth}{!}{%
\begin{tabular}{|l|l|l|l|l|l|l|}
\hline
\multicolumn{2}{|l|}{Backbone}        & ResNet10 & ResNet18 & {\bf ResNet34} & ResNet56 & ResNet101 \\ \hline
{w/ Face Model}  
                            & AEM & 6.01     & 2.98     & {\bf 1.79}     & 1.77     & 1.81      \\ 
                           {(head-pose information)}
                           & VEM & 8.07     & 4.67     & {\bf 2.87}     & 2.83     & 2.85      \\ \hline
{w/o Face Model} & AEM & 4.23     & 3.92     & {\bf 3.67}     & 3.69     & 3.65      \\  
                            {} & VEM & 6.93     & 6.52     & {\bf 5.28}     & 5.64     & 5.45      \\\hline
\end{tabular}
}
\end{center}
\vspace{-5mm}
\caption{Comparison of HGD framework with different backbone structures on In-car Gaze dataset. Despite that Resnet-56 and Resnet-101 may achieve slightly better accuracy than Resnet-34 in certain scenarios, we choose Resnet-34 as the main backbone due to its relatively much better computational efficiency.}
\vspace{-4mm}
\label{Table:_depth}
\end{table}

{\bf Single Eye vs Double Eye:} When a person is gazing at an object, both eyes have different gaze angles due to the distance between two pupils causing asymmetry. Gaze angles from both eyes should not be regarded as the same as assumed by many existing datasets \cite{zhang2015appearance, smith2013gaze, funeyediapes2014}. This assumption would potentially risk the accuracy of gaze estimation. Under this insight, during collection, we purposely collect the specific gaze ground-truth labels for both eyes independently. To the best of our knowledge, our dataset is the only dataset that directly labels the difference between right eye's and left eye’s gaze angles. Furthermore, we conduct extensive comparison experiments focusing on different means of merging both eyes during gaze estimation. These methods include: SEM, BEH, BEV and BEC. Note: SEM is the abbreviation for {\bf S}ingle {\bf E}ye {\bf M}ethod where the algorithm only takes one eye at a time and outputs one gaze; BEH is the abbreviation for {\bf B}oth {\bf E}yes to be {\bf H}orizontally stitched together and used as the input; BEV is the abbreviation for {\bf B}oth {\bf E}yes to be stitched together {\bf V}ertically (supplementary, Sec. Merging Double Eyes). Since In-car Gaze is the only one that directly keeps different gaze labels for both eyes from the same face thus these comparison experiments could only be conducted on In-car Gaze, as in Tab. 3. 

\begin{figure}[t]
\begin{center}
  \includegraphics[width=1\linewidth]{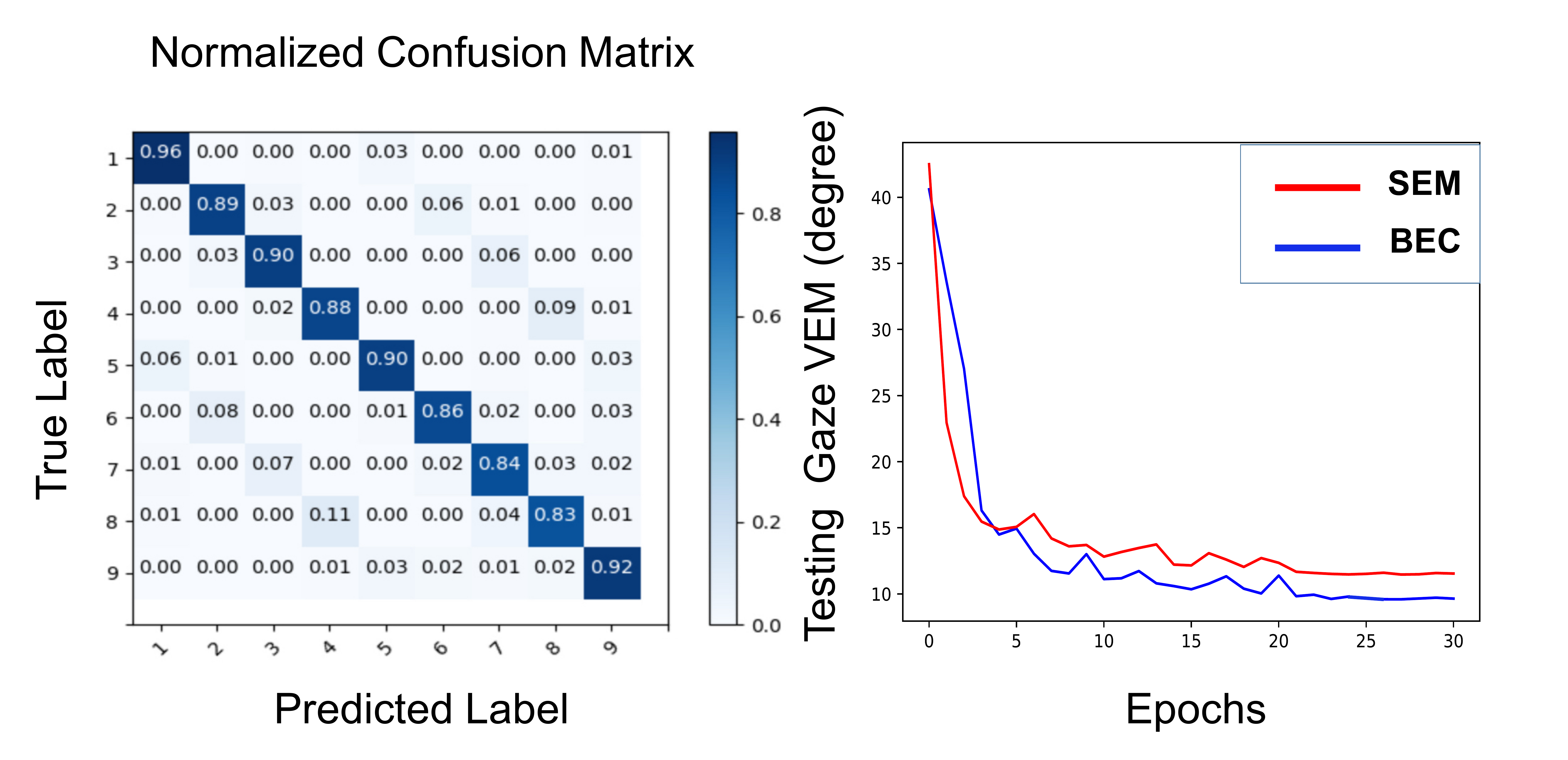}
\vspace{-4mm}
\end{center}
\vspace{-8mm}
  \caption{Left is a classification confusion matrix of HGD with explicit learning. The labeled number from 1 to 9 represents the 9 sub-spaces in front of the driver. Each sub-space represents an object, \textit{e.g.} rear mirror (supplementary, Sec. Regression and Classification). Right is the test gaze error graph for both SEM and BEC methods within 30 epochs given limited 10000 units of data in our collected dataset. Under the same settings, test gaze error of BEC decreases faster than SEM. Best viewed in color.}
\label{fig:10}
\vspace{-6mm}
\end{figure}

From the results, we conclude  that BEC help the algorithm perform the best in both accuracy and computation efficiency. BEC could potentially find the correlation between both eyes in gaze estimation during training. When we conduct the comparison experiments on our full collected dataset, we note that BEC outperforms the SEM method by around 1 degree in accuracy. Furthermore, when we limit the training dataset to only 20,000 eye images (equivalent to 20,000 input units for SEM or 10,000 input units for BEC method), BEC method outperforms the SEM by almost 2 degrees in accuracy. We plot out the test gaze error graph for both SEM and BEC methods within 30 epochs given limited data and find out that, under the same settings,  test gaze error of BEC would decrease faster than SEM, as in the right of Fig. \ref{fig:10}. We believe this is due to the relatedness between right eye and left eye. This relatedness is easier for BEC to learn given both eyes from the same face especially when the training data is limited.

Different from \cite{cheng2018appearance} focusing on the difference, asymmetry of two eyes and trying to optimize gaze prediction through the better one between two streams, our methods try to learn the difference, asymmetry, through a single stream of fewer parameters. We believe the similarity between two eyes is substantial enough for the model to learn the difference.

\vspace{-2mm}
\begin{table}[hbt]
\begin{center}
\centering
\scriptsize
\resizebox{\columnwidth}{!}{%
\begin{tabular}{|c|c|c|c|c|c|c|c|}
\hline
Dataset                            & Method      & AEM & VEM  & Dataset                                    & AEM  & VEM  & FLOPS(G) \\ \hline
{} & SEM & 2.1 & 3.44   & {} & 7.72 & 11.57 & 0.627    \\ \cline{2-4} \cline{6-8} 
                        In-car Gaze & BEH & 2.04  & 3.33  & In-car Gaze    & 7.31 & 11.08 & 0.624    \\ \cline{2-4} \cline{6-8} 
                        (full)    & BEV & 2.2 & 3.55  &     
                        (20,000 Eyes) & 7.54 & 11.36 & 0.624    \\ \cline{2-4} \cline{6-8} 
                                   & BEC & {\bf 1.79} & {\bf 2.87} &  & {\bf 6.19} & {\bf 9.42}  & {\bf 0.32}     \\ \hline
\end{tabular}
}
\end{center}
\vspace{-4mm}
\caption{Comparison of using single or double eyes. BEC concatenates both eyes on the channel level. Thus the shape of input for both eyes would change from $2 \times W \times H \times C$ to $W \times H \times 2C$. For all the methods using both eyes as inputs, the algorithms would output the gaze angles of both eyes respectively and the final error is calculated by averaging both eyes’ errors together.}
\label{Table:3}
\vspace{-5mm}
\end{table}

\vspace{-3mm}

\section{Conclusion}

\vspace{-2mm}

In this work, we fully analyze the insufficiency of current methods and datasets on incorporating head-pose information into gaze estimation. We propose our frameworks that could better incorporate head-pose into gaze estimation in two scenarios.  We further collect our own dataset to better evaluate our algorithms. Extensive evaluations demonstrate the advantage of our algorithms in free-head movement and our dataset in richer head-gaze distribution. 

\section{Acknowledgement}

The writing of this paper was partially advised by Bo Wu.

{\scriptsize
\bibliographystyle{ieee}
\bibliography{egbib}
}

\end{document}